# IRIS RECOGNITION FOR PERSONAL IDENTIFICATION USING LAMSTAR NEURAL NETWORK

Shideh Homayon[1]

Mahdi Salarian[2]

[1] University of California, Santa Cruz
City, USA

[2] University of Illinois at Chicago
Chicago, USA

**ABSTRACT**

*Iris recognition is one of the most important biometric recognition method. This is because the iris texture provides many features such as freckles, coronas, stripes, furrows, crypts, etc. Those features are unique for different people and distinguishable. Such unique features in the anatomical structure of the iris make it possible the differentiation among individuals. So during last year's huge number of people have been trying to improve its performance. In this article first different common steps for the Iris recognition system is explained. Then a special type of neural network is used for recognition part. Experimental results show high accuracy can be obtained especially when the primary steps are done well.*

## 1. INTRODUCTION

### 1.1 Biometric in general

Biometrics refers to the identification of human identity via special physiological traits. So scientists have been trying to find solution for designing technologies that can analysis those traits and ultimately distinguish between different people. Some of popular Biometric characteristic are features in fingerprint, speech, DNA, face and different part of it and hand gesture. Among those method face recognition and speaker recognition have been considered more than other during last 2 decades. The idea of automated iris recognition has been proposed firstly by Flom and Safir. They showed that Iris is an accurate and reliable code in biometric identification. Iris is the colored part of the eye behind the eyelids, and in front of the lens. Actually, it is the only internal organ of the body which is normally externally visible. It is really interesting to know this fact that those visible patterns are unique to all individuals and that the probability of finding two individuals with identical iris patterns is almost zero. Also iris pattern even for left and right eyes is different. Another fact that make identification base on Iris really powerful tools is the human iris is not changeable and is stable from one year of age until death. So the patterns of the iris are almost constant during a person's lifetime. As a result by use of a features that are highly unique the chance of having two individual having the same features is minimal. Considering those uniqueness and proposing algorithm to could extract iris correctly would lead to stable and accurate system for solving human identification problem. Although some new researches revealed there are some methods to hack this type of systems (such as capturing image form person Iris in press conference ), still iris recognition is a reliable human identification technique and reliable security recognition system. For this research we not going to capture new image by camera, instead a famous data set (CASIA database [1]) is used to evaluate results. This dataset contains thousands of different images and publicly is available upon request.

## 1.2 Background

Ophthalmologists Alphonse Bertillon and Frank Burch were one among the first to propose that iris patterns can be used for identification systems [2, 13] but John Daugman [3] was the first to design a system for the iris identification. Another valuable work proposed by R.Wildes et al. Their method was different both in the algorithm for extracting iris code and the pattern matching technique. Since the Daugman system has been shown high performance and really low failure rate, his systems are patented by the Iriscan Inc. and are also being commercially used in Iridian technologies, British Telecom, UK National Physical Lab etc. So in our research, the Daugman model is used for extracting iris pattern. Besides using common steps used in other works such as image acquisition and preprocessing, iris localization and normalization, our research utilize a powerful neural networks, say LAMSTAR [9] for recognition part. Because of availability of Daugman model [6, 7] and related source code a quick review is provided in each section to describe the theoretical approach and their results. The paper mainly focused on used neural network and its implementation along with initial experimental result and suggestion for improve of performance.

## 1.3 Image acquisition

This step is one of the most important and deciding factors for obtaining good result. A good and clear image eliminates the process of noise removal and also helps in avoiding errors in calculation. In this case, computational errors are avoided due to absence of reflections, and because the images have been taken from close proximity. This project uses the image provided by CASIA (Institute of Automation, Chinese Academy of Sciences. These images were taken for the purpose of iris recognition software research and implementation. Infra-red light was used for illuminating the eye, and so they do not have any specular reflections. So here some initial steps for decreasing error originated from reflection is not necessary. It is clear that for real-time application reflection removal process is needed.

## 2. IRIS LOCALIZATION

### 2.1 Method

The part of the eye containing information is only the iris region. As is shown iris is located between the scalera and the pupil. So it is necessary to get the iris from eye image. Actually a segmentation algorithm should be used to find the inner and outer boundaries. There are huge number of research for image segmentation such as [5] or that is based on more sophisticated algorithm but the most popular method for segmentation is edge detection. For this purpose Cany edge detector has been shown successful. The Canny detector mainly have three main steps that are finding the gradient, non-maximum suppression and the hysteresis thresholding [8, 11]. As proposed by Wildes, by considering the threshold in a vertical direction the effect of the eyelids would be decreased. Although it reduces the pixels on the circle boundary, but by utilizing Hough transform, successful localization of the boundary can be obtained even with the absence of few pixels. It is also computationally faster since the boundary pixels are lesser for calculation. Using the gradient image, the peaks are localized using non-maximum suppression. It works in the following manner. For a pixel gradian_image(x,y), in the gradient image, and given the orientation theta (x,y), the edge intersects two of its 8 connected neighbors. The point at (x, y) is a maximum if its value is not smaller than the values at the two intersection points. The next step, hysteresis thresholding, eliminates the weak edges below a low threshold, but not if they are connected to an edge above a high threshold through a chain of pixels all above the low threshold. It means the pixels above a threshold T1 should be separated. Then, these points are marked as edge points only if all its surrounding pixels are greater than another threshold T2. The threshold values were found by trial and error, and are 0.2 and 0.19 based on [8].

## 2.2 Normalization

Extracted iris has different size and value. To feed this pattern to a classifier all pattern should be normalized. For normalization of iris regions a technique based on Daugman's rubber sheet model [6,7] was employed. In this method the reference point is centre of the pupil and radial vectors pass through the iris region, as shown in Figure 1. A number of data points are selected along each radial line and this is defined as the radial resolution. The number of radial lines going around the iris region is defined as the angular resolution. Because the pupil can be non-concentric to the iris, a remapping equation should be used to rescale points depending on the angle around the circle.
This is given by

$$r' = \sqrt{\alpha\beta} \pm \sqrt{\alpha\beta^2 - \alpha - r_I^2}$$

with  (1)

$$\alpha = o_x^2 + o_y^2$$

$$\beta = \cos\left(\pi - \arctan\left(\frac{o_y}{o_x}\right) - \theta\right)$$

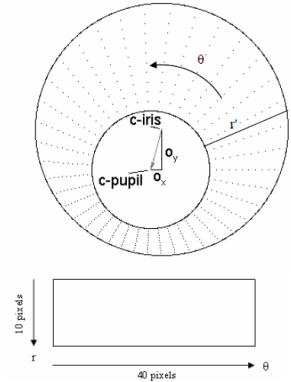

Figure 1

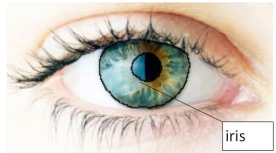

Figure 2. Result of iris localization

Here the displacement of the center of the pupil relative to the centre of the iris is given by $o_x$, $o_y$ and r' is the distance between the edge of the pupil and edge of the iris at an angle, θ around the region, and $r_I$ is the radius of the iris like Fig (1). The remapping equation first gives the radius of the iris region 'doughnut' as a function of the angle θ. A constant number of points are chosen along each radial line, then a constant number of radial data points are taken at a particular angle. The normalized pattern was made by backtracking to find the Cartesian coordinates of data points from the radial and angular position in the normalized pattern. From the 'doughnut' iris region, normalization produces a 2D array with horizontal dimensions of angular resolution and vertical dimensions of radial resolution. The result for iris localization is shown in Fig (2). In this section all the procedure is the same as [10] model including removing rotational inconsistencies that is done at the matching stage based on Daugman's rubber sheet model.

## 2.3 Results of localization AND NORMALIZATION

The normalization process proved to be successful and some results are shown in Figure 3. However, the normalization process was not able to perfectly reconstruct the same pattern from images with varying amounts of pupil dilation, since deformation of the iris results in small changes of its surface patterns. Normalization of two eye images of the same iris is shown in Figure 3.3. The pupil is smaller in the bottom image, however the normalization process is able to rescale the iris region so that it has constant dimension. In this example, the rectangular representation is constructed from 10,000 data points in each iris region. Note that rotational inconsistencies have not been accounted for by the normalization process, and the two normalized patterns are slightly misaligned in the horizontal (angular) direction. The result of whole process is shown in fig (3). For all images in the folder the template is calculated that is actually a matrix. Size of matrix is 20×480. Then those matrix are saved to be used in future as a training set. This process is shown in figure (4).

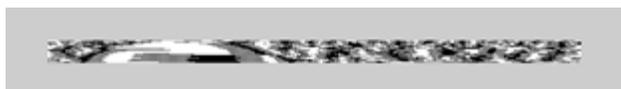

Figure 3. resulting matrix after normalization

## 4. CLASSIFIER

In order to provide accurate recognition of individuals, neural network can be used. For this research a special neural networks has been used. So after making our template and some initial steps mentioned before we have a matrix with the dimension of 20× 480. So for 16 number of class our classifier should be trained. In the next section implementation using LAMSTAR Neural network has been discussed. We decided to test it because it has been shown that is really powerful in other problems such as character recognition problem.

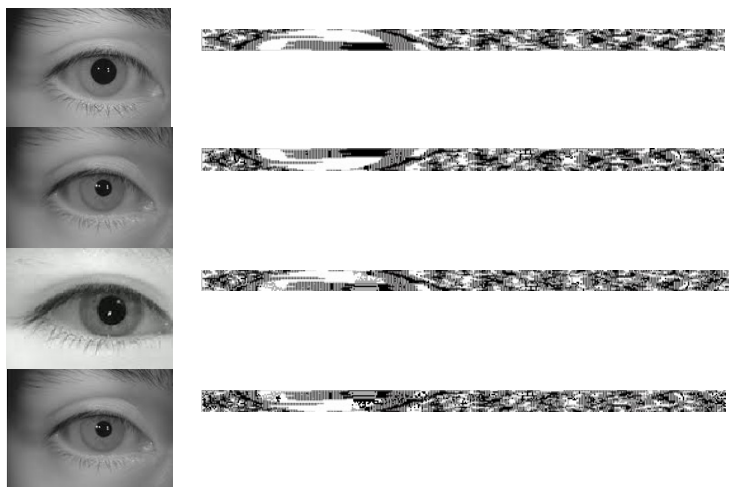

Figure 4. Training set

### 4.1 LAMSTAR neural network

### 4.1.1 Introduction to LAMSTAR

The problem consists in the realization of a LAMSTAR Artificial Neural Network for IRIS recognition. The LAMSTAR neural network, is a complex network, made by a modified version of Kohonen SOM modules. It doesn't need of the training. In fact, the input patterns are divided into many subwords, for example we considered columns of template as our subwords, so we have 480 subwords. These subwords are used for setting the weights of the SOM modules of the LAMSTAR. When a new input word is presented to the system, the LAMSTAR inspects all weights in SOM. If any pattern matches to an input subword, it is declared as winning neuron for that particularly subword. The SOM-module is based on "Winner take All" neurons, so the winning neuron has an output of 1, while all other neurons in that SOM module have zero output. Here, the SOM is built statically. This means that for every subword, we instantiate every time a new matrix that represents the SOM, and if computing the products between the stored weights and the input subword, we obtain a winner "1", we don't establish a new neuron. Otherwise, if computing that products, no one of the neurons that are present in the SOM module converge to "1", in other words, if we haven't a winner neuron, we instantiate a new neuron in the SOM module.
Every time that we instantiate a neuron, we normalize the new weights following the function such as [12, 14]:

$$x'_i = \frac{x_i}{\sqrt{\sum_j x_j^2}}$$

To converge the output of the winning neuron to "1" we follow the function below:

$$w_{(n+1)} = w_{(n)} + \alpha[X - w_{(n)}]$$

Where $\alpha = 0.8$ and it is the learning constant, w is the weight at the input of the neuron, and x the subword. A particular case could happen: when the second training pattern is input to the system, this is given to the first neuron, and if its output is close to "1", another neuron isn't built. We create neurons only when a distinct subword appears. The output layer is provided by the punishment and reward principle. If an output of the particular neuron is what is desired, the weight of the output layer is rewarded by an increment, while punishing it if the output is not what is desired.
We'll explain better this layer in the design section, reporting also the code for the sake of clarity.

### 4.1.2 Design

The design of the Neural Network is represented in the figure (5). In this network, we have 16 different representations for eyes which are both left and right eyes of 8 person. The input pattern is templates that has been extracted from images using last pre-processing steps. The size of those templates after normalization is 20× 480. Here we considered each column as a word so each word is a vector with size of 20. Also for each person 5 different images is used for training. So we selected images from data set from folders that have more than 5 images for each case to could use reminder for the testing, So after making subwords, we normalize every subword with respect itself, as we said in the introduction section. After the normalization of the input subwords, we have to train the system starting from the SOM layer. We call a function every time that we change the subword. As we can read, we initialize the som_out (which is the current SOM module), and then if we haven't a winning neuron we create it (flag=0), Other-wise we take the current neuron as winning neuron. Once that the weights of the som modules are set (w_som), we proceed to the output training. This is complex because we have to look to the sum of all the weights between the winning neurons of the SOM modules and the output layer (they are firstly set to zero). If the sum of all the weights is negative, we understand that result as "0". If is positive, we understand as "1". So the punishment and the reward is based on adding a small increment. Obviously for a negative sum, the punishment consist into adding a small positive increment, while the reward on adding a small negative increment. And vice versa for the positive sum. In this way, the system converges faster to the desired output if there's a reward, and it takes long if there's a punishment.  Briefly, the algorithm follows this few steps:

> 1)  Get the train patterns
> 2) Realize the subwords for every pattern
> 3) Normalize every subword
> 4) Set the weights of SOM module, creating every time a new neuron if it isn't a winning neuron for the new subword.
> 5) Set the output of the winning neuron to 1.
> 6) Set the weights of the Decision Layer to zero
> 7) Adjust the weights of last layer taking into account the desired output, with punishment and reward principle.

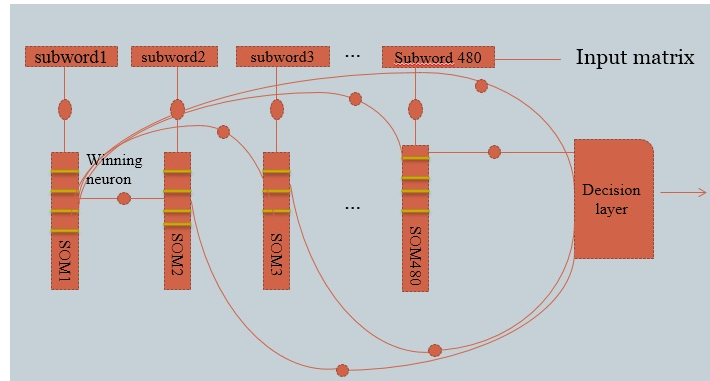

Figure 5. LAMSTAR structure

### 4.1.3) Normalized version of LAMSTAR

Based on the reward/punishment if in desire firing, a neuron is to be fired then the link weights will be rewarded. In case this happens for a couple of time the link weight value can be high enough to cause undesired neuron firing. To avoid this situation we use normalized LAMSTAR neural network in which we divide link weights by number of times the corresponding neuron was rewarded for desire firing. Considering advantages mentioned above we can add more positive points to LAMSTAR if we use the normalized version. Link weight of a neuron will not grow gradually if it wins much time. Convergence time will be reduced since normalization improve desired firing and Increase efficiency.

## 5. RESULT

The LAMSTAR and modified LAMSTAR are applied on CASIA interval database. Both of them are really fast. For instant required time for training was 66.1584s and for testing 2.5939 seconds while the accuracy was 99.39% for regular LAMSTAR and 99.57% for modified LAMSTAR. After tracing the program on each individual image I found preprocessing needs to be modified. Actually the performance of any classifier is directly depended to performance of algorithm used for finding template. For example rotational inconsistency should be taken into account. So steps including segmentation and normalization must be improved to be able to Get iris accurately and make template that is input of our neural network. It seems with having accurate templates the performance would be increased.

| Algorithm | Recognition rate |
|---|---|
| Duagman | %98.58 |
| LAMSTAR | %99.39 |
| Modified LAMSTAR | %99.57 |

Table1. Comparison between performance of our proposed method and Duagman

## 5 Conclusion and Future work:

In this work a new neural network method is presented for iris identification. A template is achieved using Image processing techniques. Classification is mainly done by LAMSTAR neural network. Structure of this network makes it a good candidate for classifying. The software code for image processing and the network has been written in MATLAB R2014a taking into account image processing toolbox and the fact that it is very user friendly in image processing application [11]. After reprocessing step all template matrix are saved and in the next step they are loaded as input to classifier. Overall result suggests that normalized LAMSTAR increase efficiency and convergence time. The next step for increasing efficiency is considering rotational inconsistency. Also it seems that having a matrix with 480 columns is not reasonable so reducing its size can be helpful especially for reducing memory that is needed for running for database with more image. In comparison with other methods the performance of Normalized LAMSTAR seems to be better and convergence time is pretty much faster than method based on other network such as Back Propagation. Also stability and not being sensitive to initialization are other positive points of using LAMSTAR. Ability to dealing with incomplete and fuzzy input data sets make LAMSTAR neural network an effective candidate for problems such as Iris classification purpose.